\documentclass{article}
\usepackage[T1]{fontenc}

\usepackage{microtype}
\usepackage{graphicx}
\usepackage{subcaption}
\usepackage{booktabs} % for professional tables
\usepackage{bm} % for bold math symbols

\usepackage{xurl}
\usepackage{hyperref}

\usepackage[preprint]{icml2027}

\usepackage{amsmath}
\usepackage{amssymb}
\usepackage{mathtools}
\usepackage{amsthm}

\usepackage[capitalize,noabbrev]{cleveref}

\theoremstyle{plain}

\theoremstyle{definition}

\theoremstyle{remark}

\usepackage[disable,textsize=tiny]{todonotes}

\icmltitlerunning{GOLeWM: Learning Object-Centric World Models with Joint Global and Slot Prediction}

\newcommand{\draftnote}[1]{}

\begin{document}

\twocolumn[
  \icmltitle{GOLeWM: Learning Object-Centric World Models \texorpdfstring{\\}{}with Joint Global and Slot Prediction}

  \begin{icmlauthorlist}
    \icmlauthor{Adriano Donninelli}{area}
    \icmlauthor{Emanuele Panizon}{area}
  \end{icmlauthorlist}
  \icmlaffiliation{area}{AREA Science Park}
  \icmlcorrespondingauthor{Adriano Donninelli}{adriano.donninelli@hotmail.it}

  \icmlkeywords{World Models, Object-Centric Learning, Visual-Goal Planning}
  \vskip 0.3in
]

\printAffiliationsAndNotice{}

\begin{abstract}
Joint-Embedding predictive architectures (JEPAs) provide an efficient framework for learning world models by predicting how latent representations change in response to actions. In many such models, the learned representation summarizes the scene, leaving individual entities and interactions implicit.
Object-centric encoders make individual objects explicit, but often rely on image or feature reconstruction during training. In this work, we introduce GOLeWM (Global and Object-centric LeWorldModel), a framework that jointly learns object-centric representations and their action-conditioned dynamics from pixels without reconstruction or pretraining.
GOLeWM trains a shared object-centric encoder with two predictive branches. The global branch aggregates object representations into a scene embedding and predicts how it changes, with SIGReg discouraging collapse. The object-level branch predicts how entities evolve, allowing planning directly in object space.
We benchmark GOLeWM on four visual-control environments, achieving a mean success rate of \textbf{92.9\%}, exceeding the averages for LeWorldModel (\textbf{79.8\%}),
Sub-JEPA (\textbf{86.1\%}), and DINO-WM (\textbf{84.8\%}).
\end{abstract}

\section{Introduction}
\label{sec:introduction}

World models support planning by predicting the consequences of
actions~\cite{hafner2019planet}. This requires visual representations
that retain information relevant to dynamics and control.
Joint-Embedding Predictive Architectures
(JEPAs)~\cite{lecun2022path,assran2023ijepa}
predict target latent representations rather than reconstructing
observations. Latent prediction alone admits a collapsed solution:
the encoder can map every observation to the same embedding, and the
predictor can match every target by returning that
embedding~\cite{balestriero2025lejepa}.

LeWorldModel (LeWM)~\cite{maes2026lewm} addresses collapse with a
simple recipe for learning world models directly from pixels: it
combines action-conditioned latent prediction with the Sketched
Isotropic Gaussian Regularizer (SIGReg), which encourages the
embeddings to follow an isotropic Gaussian distribution.

LeWM uses a single embedding for the whole scene. Human vision
organizes scenes into coherent objects from early
development~\cite{spelke1990principles}, motivating a different
approach: represent objects explicitly and model their
interactions~\cite{battaglia2018relational}. 
Object-level predictions make a plan easier to inspect: they show
how the model expects individual entities to move in response to a
proposed sequence of actions~\cite{veerapaneni2020op3}.

Most object-centric pipelines, including those based on Slot Attention,
train representations using image or feature
reconstruction~\cite{locatello2020slotattention,manasyan2025slotcontrast}.
Other object-level predictive approaches use representations learned in a
separate stage: Causal-JEPA predicts from frozen object-centric
encodings~\cite{nam2026causaljepa}, while SOLD pretrains its representations
on random-policy trajectories for model-based reinforcement
learning~\cite{mosbach2025sold}.

C-SWM jointly learns object-centric representations and dynamics without
reconstruction, using a convolutional object extractor, and evaluates
prediction in latent space~\cite{kipf2020cswm}.

We therefore ask:

\begin{itemize}
    \item \emph{Can a reconstruction-free predictive framework learn
    object-centric representations and action-conditioned dynamics jointly from pixels?}
    \item \emph{Do the learned object-centric representations focus on entities relevant to planning, and do they encode their physical state?}
\end{itemize}

To address these questions, we introduce GOLeWM, an end-to-end world
model that combines global and slot-level action-conditioned prediction
over a shared Slot Attention encoder.

\begin{figure*}[!ht]
    \centering
    \includegraphics[width=\textwidth]{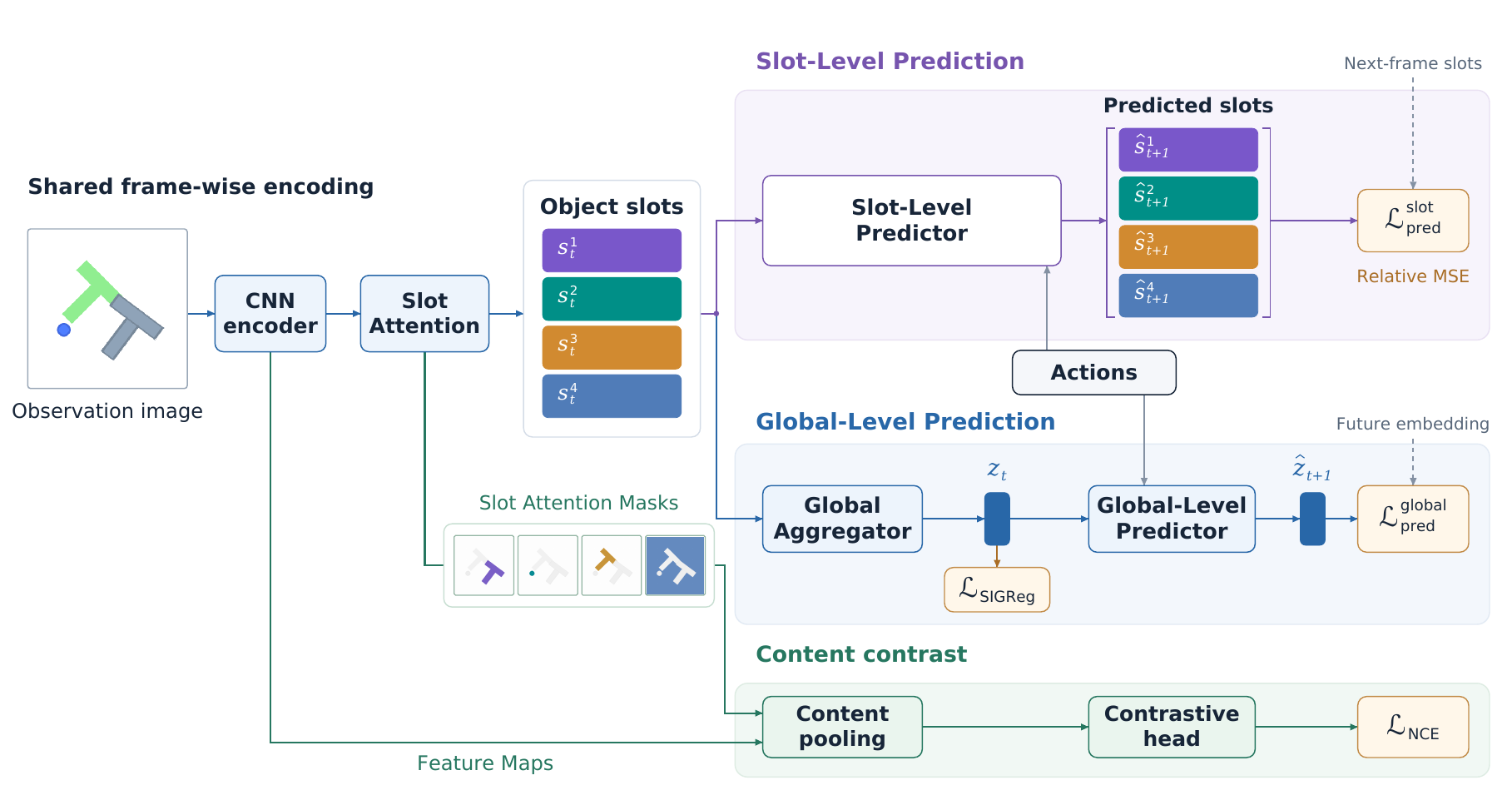}
    \caption{
        Overview of the proposed architecture.
        The visual encoder produces object slots for the global-level
        and slot-level prediction branches.
        Global embeddings are regularized with SIGReg, while content
        contrast operates on visual features pooled using Slot Attention
        masks. The slot-level predictor provides the dynamics used for
        visual-goal planning. Dashed arrows indicate prediction targets.
    }
    \label{fig:architecture}
\end{figure*}

SIGReg discourages collapse in the global embedding. Content contrast
encourages each slot index to attend to similar visual content across
frames and clips. The slot-level predictor generates rollouts for visual-goal planning.
All components are trained jointly from observation–action trajectories,
without requiring object annotations, pretraining, or reconstruction supervision.

GOLeWM achieves the highest average success across the four benchmarks
with slot-level planning (Table~\ref{tab:main-planning}). On PushT,
linear probes on the concatenated slots give the strongest position
readouts among the methods compared, and most slot variance is
concentrated in slots that track moving entities. Removing global
prediction sharply reduces PushT planning success, whereas removing
content contrast worsens the quality of attention masks.

\section{Related Work}
\label{sec:related-work}

\subsection{Joint-Embedding World Models and Regularization}
\label{sec:related-jepa}

Latent world models differ in how they learn representations and prevent collapse.
DINO-WM~\cite{zhou2025dinowm} learns action-conditioned dynamics over
frozen DINOv2 features~\cite{oquab2024dinov2}. V-JEPA learns visual
representations through feature prediction in video~\cite{bardes2024vjepa}.
V-JEPA 2-AC builds on video pretraining to learn action-conditioned
dynamics for visual-goal robot planning~\cite{assran2025vjepa2}.
PLDM~\cite{sobal2025pldm} and LeWM~\cite{maes2026lewm} jointly learn
the encoder and dynamics from pixels. LeWM combines next-embedding
prediction with SIGReg to discourage collapse.

Enforcing isotropy across the full latent space may impose a stronger
constraint than a task requires. Sub-JEPA relaxes this constraint by
applying Gaussian regularization separately in smaller random
subspaces and reports improved planning success alongside lower
effective rank~\cite{zhao2026subjepa}. Adaptive LeWorldModel uses
ordered latent prefixes and MixSIGReg to vary representational capacity
for planning~\cite{achituve2026adaptive}. \textit{No Gaussian Required}
replaces SIGReg's anti-collapse mechanism
with contrastive inverse dynamics, identifying the action associated
with a latent transition among minibatch alternatives~\cite{boylan2026nogaussian}.

\begin{figure*}[!ht]
    \centering
    \includegraphics[width=\textwidth]{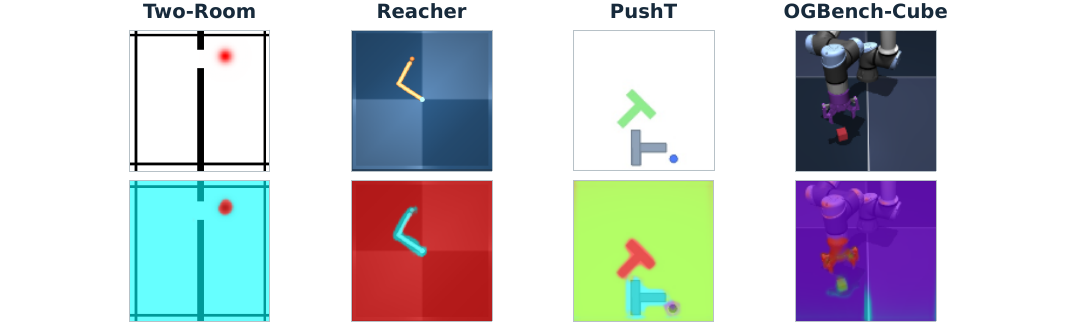}
    \caption{Example observations (top) and combined slot attention
    overlays (bottom) for GOLeWM.}
    \label{fig:slot-example}
\end{figure*}

\subsection{Object-Centric Representation Learning}
\label{sec:related-objects}

Slot Attention groups visual features into latent slots through
competitive attention, originally trained using image
reconstruction~\cite{locatello2020slotattention}.
SAVi adapts Slot Attention to video by propagating predicted slots
between frames and refining them with the current
observation~\cite{kipf2022savi}.

SlotContrast combines pretrained-feature reconstruction with slot-level temporal contrast, using cross-video negatives to discourage fixed slot-identity solutions~\cite{manasyan2025slotcontrast}.

SetCon learns slots and latent dynamics without reconstruction by contrasting aggregated predicted and encoded sets, evaluating object discovery and prediction on simple synthetic datasets rather than action-conditioned planning~\cite{loewe2020setcon}.

\subsection{Object Dynamics and Latent Planning}
\label{sec:related-dynamics}

OP3 learns object-wise dynamics for visual planning with a
reconstruction objective~\cite{veerapaneni2020op3}.
OCVP models object dynamics through factorized object-level and
temporal attention~\cite{villarcorrales2023ocvp}.
SOLD adds action conditioning for model-based reinforcement
learning~\cite{mosbach2025sold}. Its SAVi representation is pretrained
on random-policy trajectories and subsequently updated through RGB
reconstruction, while an actor and critic learn from imagined slot
rollouts.

Causal-JEPA learns object-level masked prediction over frozen,
separately trained object-centric encoders, evaluating visual reasoning
and control~\cite{nam2026causaljepa}.

Slot-JEPA~\cite{wen2026slotjepa} learns a slot-and-mask decomposition of
pretrained V-JEPA 2.1 features~\cite{murlabadia2026vjepa21} using
feature reconstruction.
The encoder and decomposition module are then frozen for
action-conditioned mask prediction and planning.
Shared reference slots establish correspondence between observations
and goals.

LPWM jointly learns particle representations and stochastic dynamics
through reconstruction-based variational training, then freezes the
world model for goal-conditioned imitation learning~\cite{daniel2026latent}.

GOLeWM combines independent frame encoding and contrast on
attention-pooled visual features with OCVP-Seq dynamics and
SOLD-style action conditioning.

\section{Method}
\label{sec:method}

GOLeWM maps each image to object slots shared by two action-conditioned
predictors (Figure~\ref{fig:architecture}). The global branch aggregates
the slots for scene-level prediction and SIGReg regularization, while
the slot branch supplies autoregressive rollouts for visual-goal planning.
Content contrast encourages consistent visual roles for the same slot
indices across frames and clips.

\subsection{Problem Setting}
\label{sec:problem-setting}

We consider an offline, reward-free setting with observation--action
trajectories collected by an expert or exploratory behavior policy.
Each trajectory contains RGB observations $\bm{o}_{1:T}$ and continuous
action blocks $\bm{a}_{1:T-1}$. We use a frame skip of five
environment steps, and $\bm{a}_t$ concatenates the intervening
actions between $\bm{o}_t$ and $\bm{o}_{t+1}$.
We learn representations and action-conditioned latent dynamics
for visual-goal planning.

\subsection{Object and Global Representations}
\label{sec:representations}

\paragraph{Object representations.}
The visual encoder in Figure~\ref{fig:architecture} processes each
observation $\bm{o}_t$ through a CNN, producing a spatial grid of
visual features. We add a 2D positional embedding and flatten the
grid into a sequence $F_t \in \mathbb{R}^{N \times d_f}$, where $N$
is the number of spatial locations and $d_f$ is the feature width.
Slot Attention~\cite{locatello2020slotattention} maps these features
to $K$ slots:
\begin{equation}
    S_t = E(\bm{o}_t) = \operatorname{SlotAttention}(F_t, Q),
\end{equation}
where $d_s$ is the slot width, $S_t\in\mathbb{R}^{K\times d_s}$,
and $Q\in\mathbb{R}^{K\times d_s}$ contains the learned initial slot
vectors after LayerNorm. The same initial vectors are used for every
observation.

We encode each image independently, without propagating slots
between frames. This allows observations and isolated goal images
to use the same encoding procedure, without an accompanying
trajectory or recurrent state. Temporal dependencies are modeled by
the causal predictors, which use only the available state and action prefixes;
during planning, their predictions are fed back autoregressively.
Figure~\ref{fig:slot-example} illustrates an example of the learned slot
assignments across the benchmark environments.

\paragraph{Global aggregator.}
A transformer aggregates each frame's slots using a learned output token.
No slot positional encodings are added, making the aggregation
invariant to permutations of the input slots.

A projection head maps the output-token representation to the global
embedding:
\begin{equation}
    \bm{z}_t = P\bigl(A(S_t)\bigr).
\end{equation}
Here $A$ returns the aggregation transformer's output-token state,
and $P$ projects it to $\bm{z}_t\in\mathbb{R}^{d_z}$.
Architectural details are provided in Appendix~\ref{app:implementation}.

\subsection{Global-Level Prediction}
\label{sec:global-prediction}

Following LeWM~\cite{maes2026lewm}, we train the global representation
using action-conditioned prediction and SIGReg~\cite{balestriero2025lejepa}, with the original regularization
weight $\lambda_{\mathrm{reg}}=0.09$.

The two prediction branches share an action encoder, producing embeddings
$\bm{e}_t=E_{\mathrm{act}}(\bm{a}_t)$ for each action block.
We train on clips of $H+1$ observations, with $H=3$. The global
predictor estimates the next embedding from each observed prefix:
\begin{equation}
    \hat{\bm{z}}_{t+1}
    = P^{\mathrm{global}}_{\phi}\bigl(\bm{z}_{1:t},\bm{e}_{1:t}\bigr),
    \qquad t=1,\ldots,H.
\end{equation}

Training uses teacher forcing, so each prefix contains encoded
observations rather than previous predictions.
A causal transformer computes these predictions in parallel. It uses
LeWM's action conditioning with a smaller, three-layer backbone
(Appendix~\ref{app:implementation}).

The global prediction loss is
\begin{equation}
    \mathcal{L}_{\mathrm{pred}}^{\mathrm{global}}
    =
    \frac{1}{H d_z}
    \sum_{t=1}^{H}
    \left\|\hat{\bm{z}}_{t+1}-\bm{z}_{t+1}\right\|_2^2,
    \label{eq:global-prediction-loss}
\end{equation}
with the average over the minibatch left implicit. SIGReg encourages
the global embeddings to follow a standard Gaussian. The global-level
branch contributes
\begin{equation}
    \mathcal{L}^{\mathrm{global}}
    =
    \mathcal{L}_{\mathrm{pred}}^{\mathrm{global}}
    + \lambda_{\mathrm{reg}}\mathcal{L}_{\mathrm{SIGReg}},
    \label{eq:global-objective}
\end{equation}
Both terms update the object-centric encoder through the global
aggregator. Gradients flow through both the context and target
embeddings in the global prediction loss.

\subsection{Slot-Level Prediction}
\label{sec:object-dynamics}

We adapt OCVP-Seq~\cite{villarcorrales2023ocvp} to predict the next
slot set from an observed slot prefix and its corresponding actions:
\begin{equation}
    \hat{S}_{t+1}
    = P^{\mathrm{slot}}_{\psi}\bigl(S_{1:t},\bm{e}_{1:t}\bigr),
    \qquad t=1,\ldots,H.
\end{equation}

The predictor follows OCVP-Seq in factorizing object interactions
and temporal dynamics into two successive attention blocks. Inspired
by SOLD's action-conditioned slot dynamics~\cite{mosbach2025sold},
we project slots and action embeddings into a shared token space and
append one action token per timestep. In each layer, the first block
attends across the slots and action token within a timestep, and the
second applies causal attention along each token's history.

The four-layer predictor uses the same sinusoidal time encoding
for all slots at each timestep. With no slot-index encodings, it is
equivariant to slot permutations. A linear projection produces a
residual update $\Delta S_t$ to the last
slot set $S_t$ in each prefix, giving $\hat{S}_{t+1}=S_t+\Delta S_t$.
Training uses the same teacher-forced observation prefixes as the
global predictor. Appendix~\ref{app:implementation} gives the
architecture settings.

\paragraph{Variance-normalized slot-level prediction.}
We train the predictor using a relative mean squared error:
\begin{equation}
    \mathcal{L}_{\mathrm{pred}}^{\mathrm{slot}}
    =
    \frac{1}{H K d_s (v_s+\epsilon_s)}
    \sum_{t=1}^{H}
    \left\|\hat{S}_{t+1}-\operatorname{sg}(S_{t+1})\right\|_F^2,
    \label{eq:slot-prediction-loss}
\end{equation}
where $\operatorname{sg}$ denotes stop-gradient,
$\|\cdot\|_F$ is the Frobenius norm, and $\epsilon_s=10^{-6}$ ensures
numerical stability. We average the loss over the minibatch using
a shared $v_s$.

Let $\sigma_{k,j}^2$ denote the variance of coordinate $j$ of slot $k$
across all context frames in the minibatch. The normalization scale
averages these variances:
\begin{equation}
    v_s =
    \operatorname{sg}\left[
        \frac{1}{K d_s}
        \sum_{k=1}^{K}\sum_{j=1}^{d_s}\sigma_{k,j}^2
    \right].
    \label{eq:slot-variance}
\end{equation}

Because slots have no direct scale constraint, we measure prediction
error relative to their empirical context variance. The loss value is
approximately invariant to uniform rescaling of predictions and
slots, provided $\epsilon_s$ is negligible relative to the variance
at both scales. Detaching the denominator
makes it a batch-dependent multiplier on the slot-loss gradient.

For slot prediction, we scale gradients reaching the encoder through
the context slots by $\alpha=0.1$ and detach the target slots.

\subsection{Contrastive Objective}
\label{sec:contrastive-objective}

Our planner compares predicted and goal slots at matching indices.
Since images are encoded independently, we use content contrast to
encourage each slot index to select similar visual content across
frames and clips.

SlotContrast~\cite{manasyan2025slotcontrast} aligns same-index slots
in adjacent frames and treats slots from other
videos as negatives, even when their indices match. This discourages
solutions based on slot identity alone. In our tasks, the same entities
recur across clips, so we treat same-index pairs from different clips as
additional positives. To limit shortcuts based on learned slot initializations,
we contrast visual features pooled under the attention masks rather
than the slot vectors themselves.

\paragraph{Attention-pooled visual content.}
Let $\bm{C}_{b,t,n}\in\mathbb{R}^{d_c}$ be the CNN feature at spatial
location $n$ in frame $t$ of minibatch clip $b$. We apply a shared
linear transform and GELU:
\begin{equation}
    \bm{h}_{b,t,n}
    = \operatorname{GELU}(W_1\bm{C}_{b,t,n}+\bm{b}_1).
    \label{eq:content-feature-transform}
\end{equation}
We pool these features with the final attention mask of slot $k$ to
obtain $\bm{c}_{b,t,k}$. We exclude boundary locations to reduce
shortcuts based on image padding and normalize the pooling weights
over the retained locations
(Eq.~\eqref{eq:attention-content-pooling}).
We then project the pooled vector to $d_q$ dimensions and apply
$\ell_2$ normalization:
\begin{equation}
    \bm{q}_{b,t,k}
    =
    \frac{W_2\bm{c}_{b,t,k}+\bm{b}_2}
         {\left\|W_2\bm{c}_{b,t,k}+\bm{b}_2\right\|_2}.
\end{equation}

\paragraph{Positive and negative pairs.}
For an anchor $a=(b,t,k)$, positives are all other
representations at slot index $k$, within and across clips.
The negative set $\mathcal{N}(a)$ contains all valid representations
at other slot indices. With temperature $\tau=0.1$, each positive
$p$ contributes a separate InfoNCE loss~\cite{oord2018cpc}:
\begin{equation}
    \ell(a,p) = \operatorname{InfoNCE}\bigl(
    \bm{q}_a,\bm{q}_p,\{\bm{q}_j\}_{j\in\mathcal{N}(a)};\tau\bigr).
\end{equation}
Each term compares one positive against the full negative set.
$\mathcal{L}_{\mathrm{NCE}}$ averages within-clip and cross-clip pairs
separately and gives the two averages equal weight.
Appendix~\ref{app:contrastive-details} gives the full loss and the
criteria for retaining features.

\subsection{Training Objective and Planning}
\label{sec:training-objective}

\paragraph{Full training objective.}
The full objective combines the global objective with slot
prediction and content contrast:
\begin{equation}
    \mathcal{L}_{\mathrm{total}}
    =
    \mathcal{L}^{\mathrm{global}}
    + w\left(
        \mathcal{L}_{\mathrm{pred}}^{\mathrm{slot}}
        + \lambda_{\mathrm{NCE}}\mathcal{L}_{\mathrm{NCE}}
    \right).
    \label{eq:full-objective}
\end{equation}
We set $\lambda_{\mathrm{NCE}}=2\times10^{-3}$ and linearly ramp $w$
from $0$ to $1$ over the first $10\%$ of training.

\paragraph{Slot-level planning.}
\label{sec:planning}

We encode the current observation and goal image as $S_t=E(\bm{o}_t)$
and $S_g=E(\bm{o}_g)$. Each candidate rollout starts from $S_t$ alone.
As the rollout advances, we append predicted slots to the context
and retain at most the most recent $H$ timesteps.
For a planning horizon of $H_{\mathrm{p}}$ action blocks, we minimize
the terminal slot cost:
\begin{equation}
    J_t\bigl(\bm{a}_{t:t+H_{\mathrm{p}}-1}\bigr)
    =
    \left\|
        \hat{S}_{t+H_{\mathrm{p}}}-S_g
    \right\|_F^2.
    \label{eq:planning-cost}
\end{equation}
Slots are compared at matching indices.
We optimize actions using the Cross-Entropy Method
(CEM)~\cite{rubinstein1999cem}. CEM repeatedly samples action sequences,
scores their predicted terminal states, and updates the sampling
distribution using the lowest-cost candidates.
For the global-level comparison, we roll out the global predictor
and compare the terminal prediction with the goal in global embedding
space, using the same CEM settings.

\section{Experiments}
\label{sec:experiments}

\subsection{Experimental Setup}
\label{sec:experimental-setup}

\paragraph{Datasets and tasks.}
We evaluate GOLeWM on the four continuous-control tasks in the
LeWM benchmark~\cite{maes2026lewm}. In Two-Room, an agent navigates
through a doorway between two rooms~\cite{sobal2025pldm}.
Reacher comes from the DeepMind Control Suite~\cite{tassa2018control};
the arm must match a target configuration. PushT is a planar
T-shaped block-pushing task~\cite{chi2023diffusionpolicy}. As in LeWM,
we use the visual-goal versions of Reacher and PushT described by
DINO-WM~\cite{zhou2025dinowm}. OGBench-Cube uses the single-cube
manipulation environment from OGBench~\cite{park2025ogbench}.

We use two slots for Two-Room and Reacher and four for PushT and
OGBench-Cube. Figure~\ref{fig:slot-example} shows observations and
an example of the learned slot assignments.

\paragraph{Planning evaluation and training runs.}
We evaluate on visual goals sampled from recorded trajectories. For
each environment, we train three independent models and evaluate
each on 300 episodes. Appendix~\ref{app:planning-evaluation} gives
the evaluation settings and aggregation protocol. We compare with
published baseline scores with matching evaluation settings.

\begin{table*}[!ht]
    \centering
    \small
    \setlength{\tabcolsep}{5pt}
    \caption{Planning success rate (\%; higher is better).
    GOLeWM and OC-RGB report mean $\pm$ sample standard deviation
    across three training runs. Published baselines and their
    uncertainties come from Table~1 of \citet{zhao2026subjepa}.
    Mean is the unweighted average across environments. Bold marks
    the highest mean in each column.}
    \label{tab:main-planning}
    \begin{tabular}{lccccc}
        \toprule
        Method & Two-Room & Reacher & PushT & OGBench-Cube & Mean \\
        \midrule
        PLDM
          & $97.0$ & $78.0$ & $78.0$ & $65.0$ & $79.5$ \\
        LeWM
          & $84.3\pm4.2$ & $82.7\pm4.4$
          & $84.7\pm6.5$ & $67.3\pm5.0$ & $79.8$ \\
        Sub-JEPA
          & $95.0\pm2.8$ & $84.0\pm4.0$
          & $89.0\pm5.3$ & $76.3\pm6.0$ & $86.1$ \\
        DINO-WM (w/o proprio.)
          & $\mathbf{100.0}$ & $79.0$
          & $74.0$ & $86.0$ & $84.8$ \\
        \midrule
        GOLeWM (global-level planning)
          & $89.8\pm0.2$ & $83.0\pm1.3$ & $88.1\pm3.4$ & $71.3\pm0.9$ & $83.1$ \\
        OC-RGB
          & $99.8\pm0.4$ & $\mathbf{84.7}\pm0.3$ & $81.0\pm3.6$ & $63.9\pm2.5$ & $82.3$ \\
        GOLeWM
          & $99.8\pm0.2$ & $83.6\pm1.2$ & $\mathbf{96.1}\pm2.5$ & $\mathbf{92.0}\pm0.9$ & $\mathbf{92.9}$ \\
        \bottomrule
    \end{tabular}
\end{table*}

\paragraph{Reconstruction-based comparison.}
\label{sec:ablation-reconstruction}
To test how well slots trained with reconstruction support
planning, we compare GOLeWM with an object-centric variant, OC-RGB. OC-RGB replaces global prediction and SIGReg with an RGB reconstruction loss,
using a slot-wise decoder that composites slot images as in SAVi~\cite{kipf2022savi}.
It retains independent frame encoding, slot prediction, and content
contrast, and uses the same slot counts and planner settings as GOLeWM.
We detach the slot predictor's context inputs ($\alpha=0$) because
training was unstable with gradients through those inputs.
Appendix~\ref{app:rgb-decoder} describes the decoder.

\subsection{Planning Results}
\label{sec:main-results}

GOLeWM with slot-level planning achieves \textbf{92.9\%} mean success,
the highest average in Table~\ref{tab:main-planning}. We follow the
evaluation protocol of \citet{zhao2026subjepa} and take the baseline
scores from their Table~1, which includes their rerun of LeWM.

GOLeWM leads the published baselines on PushT and OGBench-Cube and
comes close to the best published results on Two-Room and Reacher.
Its gains over OC-RGB are also concentrated on PushT and OGBench-Cube,
where it leads by $15.1$ and $28.1$ percentage points, respectively.
OC-RGB matches GOLeWM on Two-Room and scores slightly higher on Reacher.

Within GOLeWM, slot-level planning improves mean success by $9.8$
percentage points over global-level planning, with the largest gain
on OGBench-Cube ($20.7$ points). GOLeWM's global planner also exceeds
the LeWM reference average.

\section{Analysis of Learned Slot Representations}
\label{sec:analysis}

To understand what the planner has available in slot space, we ask
which parts of the scene slots attend to, which physical variables
their latents encode, and how variation is distributed across the
global embedding and the slots.

\subsection{Object Decomposition Across Training Runs}
\label{sec:object-analysis}

\begin{figure*}[!ht]
    \centering
    \includegraphics[width=\textwidth]{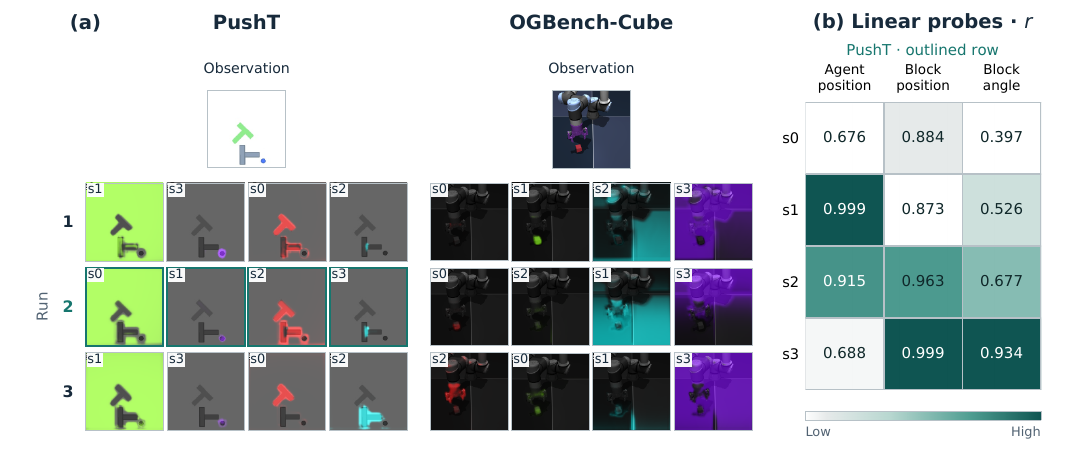}
    \caption{Slot Attention masks and linear state probes.
\textbf{(a)} PushT and OGBench-Cube observations with individual-slot
overlays from three training runs. Columns align approximate visual
roles, and labels show the original slot indices.
\textbf{(b)} Linear-probe Pearson $r$ for the slots in the outlined
PushT row of (a), in the same order. Position scores average the
correlations for
$x$ and $y$. Colors are scaled separately for each target.}
    \label{fig:cross-seed-binding}
    \label{fig:pusht-probes-slots}
\end{figure*}

We compare attention maps from three independently trained models
on the same observations. Figure~\ref{fig:cross-seed-binding}(a)
shows individual slots for PushT and OGBench-Cube.
Figure~\ref{fig:cross-seed-binding-full} extends the comparison across
frames, and Appendix~\ref{app:cross-seed-binding} shows Two-Room and
Reacher.

Across the displayed PushT sequence, every run has a slot that
highlights the pusher. Block-focused masks range from a small patch on
the block to broader coverage of its footprint. In OGBench-Cube, all
three runs highlight the cube, while coverage of the gripper, arm,
and background varies (Figure~\ref{fig:cross-seed-binding-full}).

With three training runs and near-ceiling success on some tasks,
these examples do not establish whether attention quality correlates
with planning success.

\subsection{Physical-State Probes}
\label{sec:binding-probes}

We fit linear probes to frozen global embeddings, concatenated
slots, and individual slots on PushT to measure whether position
and orientation are readable and how that information is distributed.
\Cref{fig:pusht-probes-slots}(b) shows per-slot probe results, directly comparable
with the slot attention binding masks in \Cref{fig:cross-seed-binding}(a) center row.
\Cref{tab:pusht-probes-linear} complements the analysis with the global embedding and concatenated slot results.

We report Pearson $r$ on a held-out frame split;
ground-truth state is used only for the probes.
Appendix~\ref{app:analysis-details} gives the protocol, and
Table~\ref{tab:pusht-probes-full} reports Pearson $r$ for linear and
MLP probes across all three training runs.

\begin{table}[!ht]
    \centering
    \small
    \setlength{\tabcolsep}{4pt}
    \caption{Linear-probe Pearson correlations on PushT. Position
    scores average horizontal and vertical correlations. DINO-WM is
    from Table~1 of \citet{maes2026lewm}; the other baselines are from
    Table~4 of \citet{zhao2026subjepa}. GOLeWM results use the model
    in Figure~\ref{fig:pusht-probes-slots}(b).
    Bold marks the highest displayed score in each column, including ties.}
    \label{tab:pusht-probes-linear}
    \begin{tabular}{@{}lrrr@{}}
        \toprule
        Model / representation & \shortstack{Agent\\position}
          & \shortstack{Block\\position} & \shortstack{Block\\angle} \\
        \midrule
        PLDM               & 0.955 & 0.938 & 0.745 \\
        LeWM               & 0.974 & 0.986 & 0.902 \\
        Sub-JEPA           & 0.976 & 0.988 & 0.884 \\
        DINO-WM            & 0.977 & 0.997 & \textbf{0.979} \\
        \midrule
        \multicolumn{4}{@{}l}{\textit{GOLeWM}} \\
        Global embedding   & 0.983 & 0.992 & 0.850 \\
        Concatenated slots & \textbf{0.999} & \textbf{0.999} & 0.939 \\
        Slot 0             & 0.676 & 0.884 & 0.397 \\
        Slot 1             & \textbf{0.999} & 0.873 & 0.526 \\
        Slot 2             & 0.915 & 0.963 & 0.677 \\
        Slot 3             & 0.688 & \textbf{0.999} & 0.934 \\
        \bottomrule
    \end{tabular}
\end{table}

\paragraph{Physical-state accessibility.}
The concatenated slots yield higher linear-probe correlations
for agent and block position than the published reference scores, while DINO-WM leads on block angle
($r=0.979$; Table~\ref{tab:pusht-probes-linear}). Within GOLeWM,
concatenated slots outperform the global embedding on all three
measures in every training run (Table~\ref{tab:pusht-probes-full}).
In the outlined run, the largest gain is for block angle, from
$r=0.850$ to $0.939$. This suggests that some physical state is less
accessible once slot representations are aggregated.

In the outlined PushT row of \Cref{fig:cross-seed-binding}(a), the
pusher- and block-focused slots give the strongest individual
readouts for their respective objects: Slot~1 for pusher position
($r=0.999$) and Slot~3 for block position ($r=0.999$) and angle
($r=0.934$). Block position is also accessible from Slot~2, whose
mask covers part of the block ($r=0.963$).

\subsection{Effective Representation Dimensionality}
\label{sec:dimension-analysis}

SIGReg encourages the global embedding to follow an isotropic
Gaussian distribution. Since SIGReg is applied after aggregation,
we compare how much the global embedding and the slots vary and
how that variation is distributed.

For each representation, we stack the features of the same $m$ PushT
observations into a matrix $X\in\mathbb{R}^{m\times D}$. We center each
coordinate without rescaling it, obtaining $X_c$, and compute the
sample covariance $\Sigma_X=X_c^\top X_c/(m-1)$. Its eigenvalues
$\mu_1,\ldots,\mu_D$ give the variance along each principal direction.
Appendix~\ref{app:analysis-details} gives the sampling protocol.

We first report the total variance, $\operatorname{tr}(\Sigma_X)$.
To measure how broadly that variance is spread across principal
directions, we use the covariance-based effective rank adopted by
Sub-JEPA~\cite{zhao2026subjepa}. For nonzero total variance,
\begin{equation}
    \begin{aligned}
        p_j &= \frac{\mu_j}{\sum_{i=1}^{D}\mu_i}, \\
        r_{\mathrm{eff}}(\Sigma_X)
            &= \exp\left(-\sum_{j=1}^{D} p_j\log p_j\right).
    \end{aligned}
    \label{eq:entropy-effective-rank}
\end{equation}

The global embedding has substantially higher covariance effective rank
than the concatenated slots despite its smaller dimension (Table~\ref{tab:pusht-effective-rank}).
Global variation is therefore spread more broadly across principal
directions, whereas slot variation is more concentrated.
This is consistent with the nonlinear global readout organizing
the representation to accommodate SIGReg, which is applied after
aggregation rather than directly to the slots.

For the PushT model in Table~\ref{tab:pusht-effective-rank}, the
pusher- and block-associated slots (1 and 3) account for $95.9\%$
of the concatenated representation's variance. The
background-associated slot 0 contributes less than $0.01\%$.
Across the three PushT runs, global covariance effective rank ranges
from $81.6$ to $82.7$, compared with $9.5$ to $11.5$ for the slot
concatenation. Two-Room shows the same ordering in all three runs
(Appendix~\ref{app:tworoom-diagnostics}).

\begin{table}[!ht]
    \centering
    \small
    \setlength{\tabcolsep}{3pt}
    \caption{Total variance and covariance effective rank for the PushT
    representations analyzed in Figure~\ref{fig:pusht-probes-slots}.
    $D$ denotes the embedding dimension.}
    \label{tab:pusht-effective-rank}
    \begin{tabular}{@{}lrrr@{}}
        \toprule
        Representation & $D$ & Total variance & $r_{\mathrm{eff}}$ \\
        \midrule
        Global embedding   & 192 & 191.78 & 81.67 \\
        Concatenated slots & 512 & 166.51 & 10.59 \\
        Slot 0             & 128 &   0.01 &  5.10 \\
        Slot 1             & 128 &  58.86 &  5.34 \\
        Slot 2             & 128 &   6.81 &  2.98 \\
        Slot 3             & 128 & 100.84 &  7.62 \\
        \bottomrule
    \end{tabular}
\end{table}

\section{Ablations of the Training Formulation}
\label{sec:ablation-global}
\label{sec:ablation-binding}

Table~\ref{tab:objective-ablations} reports one training run per
four-slot PushT variant. Removing global prediction leaves SIGReg
active, and removing SIGReg leaves global prediction active.
Context-slot gradients from prediction are
either scaled by $\alpha=0.1$ or detached ($\alpha=0$).

\begin{figure}[!ht]
    \centering
    \includegraphics[width=\linewidth]{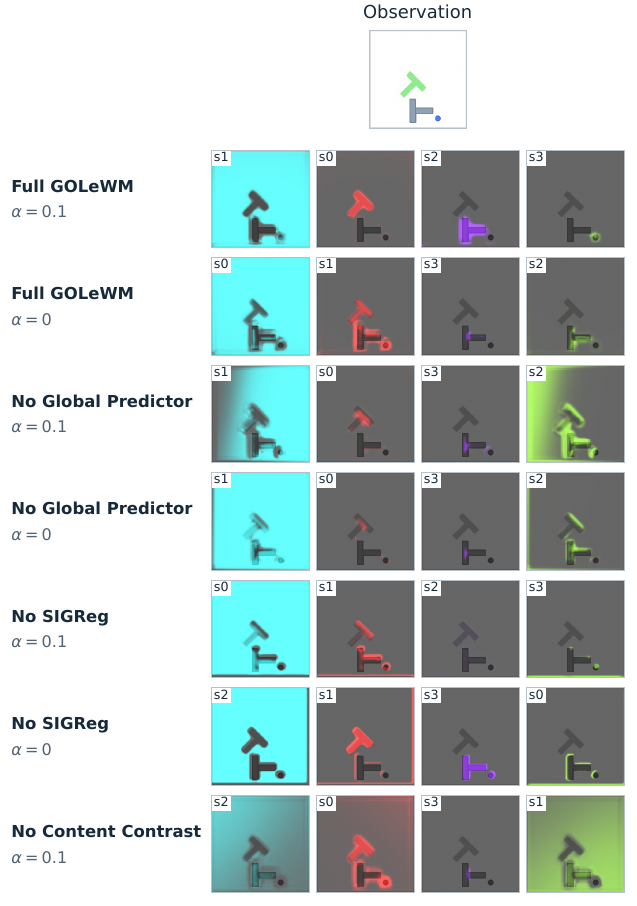}
    \caption{Slot Attention across the PushT training ablations.
    Each row shows four per-slot overlays on the shared observation
    (top). All variants use the same training seed, and columns align
    slots by approximate visual role.}
    \label{fig:pusht-ablation-bindings}
\end{figure}

With the full objective, detaching gradients through the context slots
changes planning success from $98.3\%$ to $97.7\%$ and produces less
distinct slot masks in the example shown in
\Cref{fig:pusht-ablation-bindings}. Removing the global predictor
reduces success to $56.0\%$ with scaled context gradients
($\alpha=0.1$) and $25.7\%$ with detached context gradients
($\alpha=0$). Without SIGReg, the corresponding scores are $1.3\%$
and $16.0\%$.

PushT planning success remains high without content contrast
($97.0\%$ versus $98.3\%$ with it; Table~\ref{tab:objective-ablations}).
In \Cref{fig:pusht-ablation-bindings}, removing content contrast makes
the masks overlap more and weakens the separation between foreground
and background.

\begin{table}[!ht]
    \centering
    \small
    \setlength{\tabcolsep}{3pt}
    \caption{PushT ablations. Values are mean $\pm$ standard deviation
    across six sets of evaluation episodes for each variant.}
    \label{tab:objective-ablations}
    \begin{tabular}{@{}lcc@{}}
        \toprule
        Variant & $\alpha$ & Success (\%) \\
        \midrule
        Full GOLeWM & 0.1 & $98.3\pm3.2$ \\
        Full GOLeWM & 0 & $97.7\pm2.3$ \\
        No Global Predictor & 0.1 & $56.0\pm6.8$ \\
        No Global Predictor & 0 & $25.7\pm5.4$ \\
        No SIGReg & 0.1 & $1.3\pm2.1$ \\
        No SIGReg & 0 & $16.0\pm4.2$ \\
        No Content Contrast & 0.1 & $97.0\pm2.8$ \\
        \bottomrule
    \end{tabular}
\end{table}

\section{Discussion and Conclusion}
\label{sec:conclusion}

In this work, we introduced GOLeWM and showed that object-centric
representations and action-conditioned dynamics can be learned
jointly from pixels without reconstruction or pretraining. Across
the evaluated environments, GOLeWM achieved the highest average
success when planning in slot space, outperforming both the
baselines and planning in its own global embedding. This is a first
step toward world models that learn what to track and how it moves
from the same experience.

In PushT, the concatenated slots yielded better linear readouts
than the global embedding. They also had lower effective
dimensionality, with most of their variance concentrated in slots
that tracked moving entities.

Comparing the PushT attention masks with the linear probes showed
that a slot can focus on one object while still encoding information
about another: even the pusher-focused slot carried information about
block position.

We evaluated GOLeWM with a fixed slot count per task and goals
drawn from recorded trajectories. A next step is to extend it to
scenes where objects become occluded or vary in number and identity
across episodes. Explicit matching between predicted and goal slots
could relax the planner's reliance on matching indices, while a
prediction loss that accounts for occlusion could help preserve
object identities over time.

To encourage specialization, another direction is to replace SIGReg
on the global embedding with a regularizer applied directly to the
slots. The challenge is to prevent collapse while preserving the
information the slots collectively need for prediction and planning.
For visually richer scenes, pretrained features could also help the
slot encoder discover objects, as demonstrated in real-world video
by VideoSAUR~\cite{zadaianchuk2023videosaur}.

\bibliography{references}
\bibliographystyle{icml2027}

\newpage
\appendix
\onecolumn

\section{Model and Training Details}
\label{app:implementation}

\subsection{Visual Inputs and Architecture}
GOLeWM uses $192\times192$ RGB inputs for training and planning.
At evaluation, the environment renders $224\times224$ images, which
are resized to $192\times192$ before encoding. The published LeWM and
Sub-JEPA baselines use $224\times224$ model inputs
~\cite{maes2026lewm,zhao2026subjepa}.

Table~\ref{tab:architecture-settings} summarizes the GOLeWM architecture.
We use Xavier initialization for the learned slot vectors and apply
LayerNorm before the first Slot Attention update. The slot predictor's
residual head initially outputs zero, so the predictor starts from $\hat{S}_{t+1}=S_t$.

\begin{table}[!ht]
    \centering
    \small
    \setlength{\tabcolsep}{4pt}
    \caption{GOLeWM architecture by component and module.
    Linear and convolutional dimensions denote input and output widths;
    attention settings specify heads and dimension per head where applicable.}
    \label{tab:architecture-settings}
    \begin{tabular}{@{}p{0.19\linewidth}p{0.27\linewidth}p{0.48\linewidth}@{}}
        \toprule
        Component & Module & Settings \\
        \midrule
        CNN encoder & Convolution stack & Four $5\times5$ convolutions with
            channel widths $3\to32\to64\to64\to64$, strides $(1,2,1,1)$, and ReLU after the first three. \\
         & Position embedding & Learned linear projection of a coordinate grid \\
        \midrule
        Slot Attention & Input transform & LayerNorm; Linear $64\to64$ \\
         & Slot initialization & Learned per index; LayerNorm; width 128 \\
         & Iterative updates & 3 iterations; GRU; residual MLP width 128 \\
        \midrule
        Global aggregator & Slot projection & Linear $128\to192$ \\
         & Learned tokens & 1 output token; 1 register token \\
         & Transformer $\times3$ & Pre-RMSNorm; 8 heads; SiLU MLP width 512 \\
        \midrule
        Global readout & Hidden projection & Linear $192\to1024$; BatchNorm; GELU \\
         & Output projection & Linear $1024\to192$ \\
        \midrule
        Global predictor & Causal transformer $\times3$ & 8 heads of width 64; MLP width 1024; dropout 0.1 \\
         & Temporal positions & Learned embeddings \\
         & Action conditioning & AdaLN-zero \\
        \midrule
        Shared action encoder & Conv1d & Pointwise; stacked action block to 10 channels \\
         & MLP & Linear to 768; SiLU; Linear $768\to192$ \\
        \midrule
        Slot predictor & Attention layers $\times4$ & Object attention followed by causal temporal attention \\
         & Tokens and heads & Width 384; 6 heads of width 64 \\
         & Normalization and MLP & Pre-normalization; MLP width 768; dropout 0 \\
         & Residual output & Zero at initialization; added to the latest input slots \\
        \midrule
        Contrastive head & Before pooling & Linear $64\to64$; GELU \\
         & After pooling & Linear $64\to32$; $\ell_2$ normalization \\
        \bottomrule
    \end{tabular}
\end{table}

\subsection{Training Schedule}
We use AdamW with
learning rate $5\times10^{-5}$, weight decay $10^{-3}$, and batch size
128 on one GPU with bfloat16 mixed precision. Training runs for 10
epochs, with gradient clipping at 1.0, linear learning-rate warmup
over the first $1\%$ of steps, and then per-step cosine decay to zero.
We use the final checkpoint for evaluation.

\subsection{Content Pooling and Contrastive Aggregation}
\label{app:contrastive-details}
\label{app:objectives}
This subsection specifies how the content-contrast objective in
Section~\ref{sec:contrastive-objective} pools features and weights
within-clip and cross-clip pairs.

We take $\bm C_{b,t,n}\in\mathbb{R}^{d_c}$ from the second CNN block
after ReLU and before positional embeddings, on the same $96\times96$
grid as the competition masks.
The binary mask $V_n$ excludes cells with receptive fields that
intersect padding in the full CNN, followed by one cell of erosion
($82\times82$ cells retained). For minibatch clip $b$, frame $t$,
and slot $k$, let $M_{b,t,k,n}$ be the final competition weight at
grid location $n$. We pool the transformed features from
Eq.~\eqref{eq:content-feature-transform} as
\begin{equation}
    \bm{c}_{b,t,k}
    =
    \frac{\sum_{n=1}^{N}V_n M_{b,t,k,n}\bm{h}_{b,t,n}}
         {\sum_{n=1}^{N}V_n M_{b,t,k,n}+\epsilon_c},
    \label{eq:attention-content-pooling}
\end{equation}
where $\epsilon_c$ stabilizes the denominator.
We exclude representations with no retained mask mass.

Let $a=(b,t,k)$ be an anchor and $p$ a distinct positive with the
same slot index. The negative set $\mathcal{N}(a)$ contains all
valid minibatch representations with slot index different from $k$.
Each ordered pair contributes
\begin{equation}
    \ell(a,p)
    =
    -\log
    \frac{
        \exp(\bm{q}_a^\top \bm{q}_p/\tau)
    }{
        \exp(\bm{q}_a^\top \bm{q}_p/\tau)
        +
        \sum_{j\in\mathcal{N}(a)}
        \exp(\bm{q}_a^\top \bm{q}_j/\tau)
    },
    \label{eq:content-infonce}
\end{equation}
with temperature $\tau=0.1$. Other positives are excluded from the
denominator. Let $\mathcal{E}_{\mathrm{within}}$ contain the valid
pairs from different frames in the same clip, and
$\mathcal{E}_{\mathrm{cross}}$ the valid pairs from distinct clips
(which may come from the same episode). In the objective below, omit
the term for a pair group if it is empty.
\begin{align}
    \mathcal{L}_{\mathrm{NCE}}
    &= \frac{1}{2|\mathcal{E}_{\mathrm{within}}|}
       \sum_{(a,p)\in\mathcal{E}_{\mathrm{within}}}\ell(a,p)
       \nonumber\\
    &\quad+\frac{1}{2|\mathcal{E}_{\mathrm{cross}}|}
       \sum_{(a,p)\in\mathcal{E}_{\mathrm{cross}}}\ell(a,p).
    \label{eq:content-contrast}
\end{align}

\subsection{OC-RGB Baseline}
\label{app:rgb-decoder}
OC-RGB replaces global prediction and SIGReg with an image
reconstruction objective while retaining slot prediction and content
contrast. The spatial-broadcast decoder~\cite{watters2019spatialbroadcast}
replicates each slot vector across a spatial grid and supplies position
information to a shared CNN. The CNN predicts an RGB image and a mask
for each slot at $64\times64$ resolution. At each pixel, we normalize
the masks across slots and use them to weight the sum of the slot
images. The reconstruction loss is the pixelwise mean squared error,
weighted by $100$.

We detach both context and target slots in the slot-prediction loss.
The visual encoder therefore learns from image reconstruction and
content contrast, while the slot predictor learns from latent
prediction error.

\section{Planning Evaluation and Results by Run}
\label{app:full-results}

\subsection{Evaluation Protocol}
\label{app:planning-evaluation}

We use the same four planning benchmarks as Sub-JEPA~\cite{zhao2026subjepa}.
For each episode, we sample an initial state from a training trajectory and use the observation 25 environment
steps later in that trajectory as the goal image. The planner receives
only images and actions and starts each rollout from one observed frame.
An episode succeeds if it meets the environment's goal-reaching criterion
within 50 environment steps.

For each trained model, we evaluate six sets of 50 episodes. The main
table reports the mean and sample standard deviation across three
training-run averages. The tables for individual runs and ablations
report variability across the six evaluation sets.

CEM uses 300 candidates, 30 elites, 30 iterations, and an initial
variance scale of 1.0. The horizon comprises five blocks of five
consecutive actions (frame skip 5), covering 25 environment steps.
All five blocks are executed before replanning. These settings are
shared by both planning variants and the reconstruction-based comparison.

\subsection{Planning Results by Training Run}
Tables~\ref{tab:golewm-training-seeds} and~\ref{tab:rgb-training-seeds}
break down the averages in Table~\ref{tab:main-planning} by training
run. The GOLeWM table compares the two planning spaces on the same
trained model for each run. The OC-RGB table reports slot-level
planning for the reconstruction baseline. Run numbers distinguish
repeats within each task and model variant.

\begin{table}[!ht]
    \centering
    \small
    \setlength{\tabcolsep}{5pt}
    \caption{GOLeWM planning success (\%) by training run and planning
    space. Entries are mean $\pm$ standard deviation across six sets
    of 50 evaluation episodes. Table~\ref{tab:main-planning} reports
    variability across the three training-run averages.}
    \label{tab:golewm-training-seeds}
    \begin{tabular}{llcccc}
        \toprule
        Planning space & Run & Two-Room & Reacher & PushT & OGBench-Cube \\
        \midrule
        Slot-level & 1 & $99.7\pm0.8$ & $82.3\pm2.7$ & $96.7\pm1.6$ & $93.0\pm3.3$ \\
         & 2 & $99.7\pm0.8$ & $84.7\pm5.6$ & $93.3\pm5.3$ & $91.7\pm2.7$ \\
         & 3 & $100.0\pm0.0$ & $83.7\pm5.1$ & $98.3\pm3.2$ & $91.3\pm1.6$ \\
        \midrule
        Global-level & 1 & $89.7\pm3.9$ & $81.7\pm8.0$ & $87.7\pm5.0$ & $70.3\pm5.3$ \\
         & 2 & $90.0\pm5.1$ & $84.3\pm6.3$ & $85.0\pm7.2$ & $72.0\pm6.3$ \\
         & 3 & $89.7\pm3.7$ & $83.0\pm4.3$ & $91.7\pm3.9$ & $71.7\pm4.6$ \\
        \bottomrule
    \end{tabular}
\end{table}

\begin{table}[!ht]
    \centering
    \small
    \setlength{\tabcolsep}{5pt}
    \caption{OC-RGB planning success (\%) by training run. Entries are
    mean $\pm$ standard deviation across six sets of 50 evaluation
    episodes. Slot counts match those of GOLeWM in
    Table~\ref{tab:main-planning}.}
    \label{tab:rgb-training-seeds}
    \begin{tabular}{lcccc}
        \toprule
        Run & Two-Room & Reacher & PushT & OGBench-Cube \\
        \midrule
        1 & $100.0\pm0.0$ & $84.7\pm4.3$ & $85.0\pm6.3$ & $66.7\pm6.3$ \\
        2 & $99.3\pm1.0$ & $84.3\pm6.1$ & $80.0\pm4.6$ & $62.0\pm8.9$ \\
        3 & $100.0\pm0.0$ & $85.0\pm3.5$ & $78.0\pm6.2$ & $63.0\pm7.2$ \\
        \bottomrule
    \end{tabular}
\end{table}

\section{Representation Analysis}
\label{app:representation-analysis}

\subsection{Attention Across Runs on PushT and OGBench-Cube}
\begin{figure}[!ht]
    \centering
    \includegraphics[width=\textwidth]{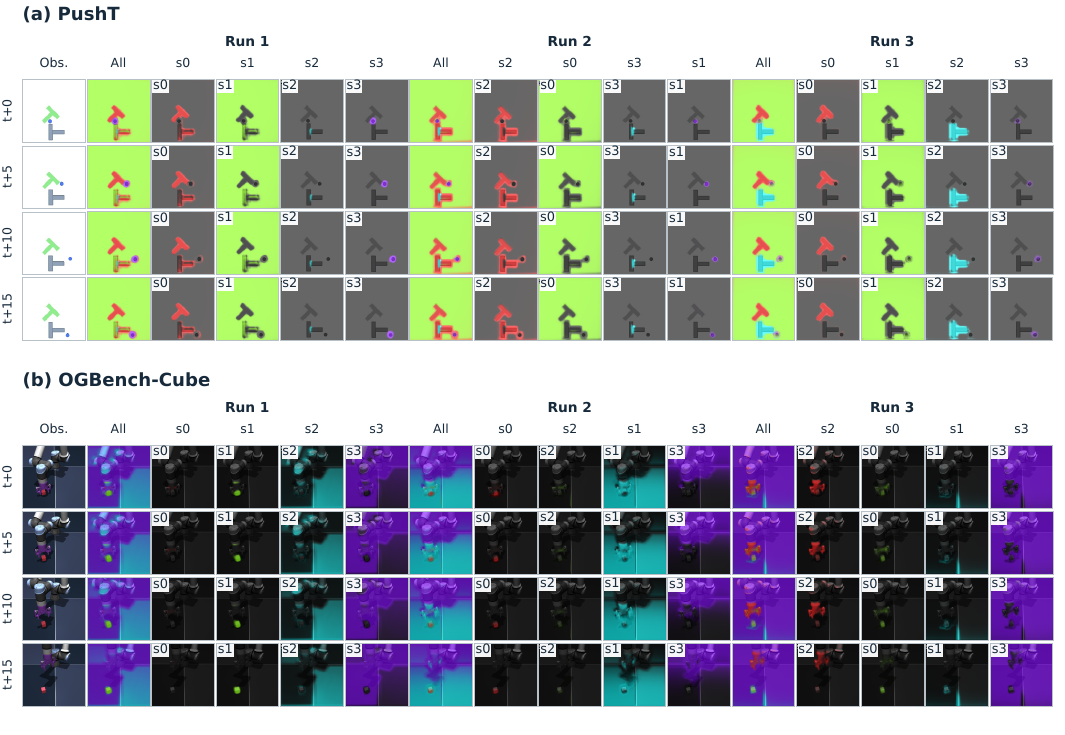}
    \caption{Combined (All) and per-slot attention overlays across
    four observations on \textbf{(a)} PushT and \textbf{(b)} OGBench-Cube.
    Slot labels retain the original indices.}
    \label{fig:cross-seed-binding-full}
\end{figure}

Figure~\ref{fig:cross-seed-binding-full} extends
Figure~\ref{fig:cross-seed-binding} to a short sequence. The same
observations are shown for all three runs within each task, so the overlays
can be compared both across frames within a run and across runs at a
fixed frame.

\subsection{Two-Room and Reacher}

\label{app:cross-seed-binding}
\begin{figure}[!ht]
    \centering
    \includegraphics[width=0.60\textwidth]{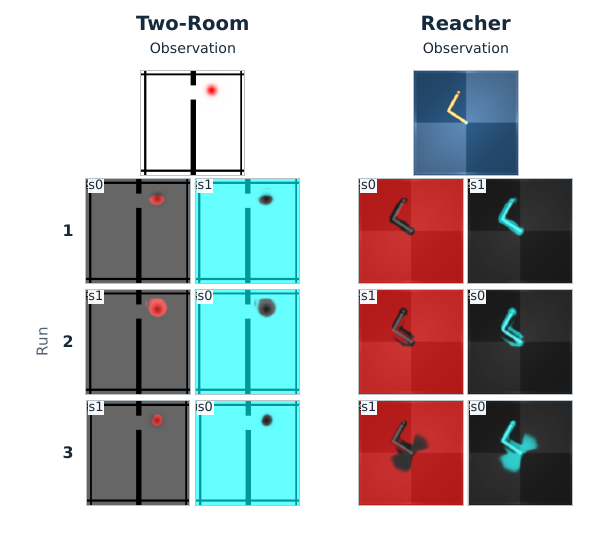}
    \caption{Slot Attention on Two-Room (left) and Reacher (right).
    Reference observations appear above the individual-slot overlays.
    Columns align approximate visual roles; labels show original slot
    indices.}
    \label{fig:cross-seed-binding-appendix}
\end{figure}

The single-frame overlays in Figure~\ref{fig:cross-seed-binding-appendix}
complement the PushT and OGBench-Cube sequences. They show how attention
is divided between the agent or arm and the rest of the scene in
Two-Room and Reacher across training runs.

\paragraph{Two-Room diagnostics.}

\label{app:tworoom-diagnostics}
Table~\ref{tab:tworoom-diagnostics} complements the attention overlays
by measuring how readily agent position can be decoded and how
variance is distributed across representations. Both the global
embedding and the concatenated slots support accurate position
readout. The global embedding has higher covariance effective rank,
while the slot associated with the agent accounts for more than
$99.9\%$ of the concatenated-slot variance in all three runs.

\begin{table}[!ht]
    \centering
    \small
    \setlength{\tabcolsep}{4pt}
    \caption{Two-Room representation diagnostics. Agent-position Pearson
    $r$ is computed with linear probes and averaged across the two
    coordinates, then rounded to three decimal places. $D$ is the embedding dimension.}
    \label{tab:tworoom-diagnostics}
    \begin{tabular}{@{}rlrrrr@{}}
        \toprule
        Run & Representation & $D$ & Agent position $r$ & $r_{\mathrm{eff}}$ & Total variance \\
        \midrule
        1 & Global embedding & 192 & 0.999 & 45.89 & 197.69 \\
         & Concatenated slots & 256 & 1.000 & 8.74 & 44.47 \\
         & Slot 0 & 128 & 1.000 & 8.74 & 44.46 \\
         & Slot 1 & 128 & 0.794 & 2.78 & $<0.01$ \\
        \midrule
        2 & Global embedding & 192 & 0.999 & 44.58 & 197.76 \\
         & Concatenated slots & 256 & 0.999 & 10.39 & 51.40 \\
         & Slot 0 & 128 & 0.876 & 3.95 & $<0.01$ \\
         & Slot 1 & 128 & 0.999 & 10.39 & 51.39 \\
        \midrule
        3 & Global embedding & 192 & 0.999 & 47.40 & 197.54 \\
         & Concatenated slots & 256 & 1.000 & 9.70 & 48.65 \\
         & Slot 0 & 128 & 0.950 & 3.58 & $<0.01$ \\
         & Slot 1 & 128 & 1.000 & 9.70 & 48.65 \\
        \bottomrule
    \end{tabular}
\end{table}

\subsection{PushT Representation Diagnostics}
\label{app:analysis-details}

For each PushT checkpoint, we sample 4,000 frames uniformly without
replacement from the world-model training dataset. We extract frozen
representations for both analyses from these frames.

\paragraph{Total variance and effective rank.}
We estimate the covariance of each representation using all sampled
frames, following Section~\ref{sec:dimension-analysis}.

\paragraph{Physical-state probes.}
We split the sample into 2,800 probe-training frames and 1,200 test
frames. The linear probes use ridge regression with a penalty of 1.0
on the raw latents. The MLP probes use one 256-unit GELU hidden layer
and standardize inputs and targets using training statistics. We reserve
420 of the probe-training frames for MLP validation. We train the MLP
with AdamW for up to 200 epochs (batch size 128, learning rate $10^{-3}$,
weight decay $10^{-2}$), using a patience of 40 epochs and restoring
the best validation checkpoint.

For each position, we average the Pearson correlations for the two
coordinates. For orientation, we use the dataset's scalar block angle.
Table~\ref{tab:pusht-probes-full} reports both probe types for all
three training runs.

\begin{table}[!ht]
    \centering
    \small
    \setlength{\tabcolsep}{6pt}
    \caption{Pearson correlations ($r$) for linear and MLP probes of
    PushT state across three GOLeWM training runs. Position scores
    average the correlations for the two coordinates. Within each run,
    bold marks the highest score among individual slots for each target
    and probe type.}
    \label{tab:pusht-probes-full}
    \begin{tabular}{@{}l*{3}{rr}@{}}
        \toprule
        & \multicolumn{2}{c}{Agent position}
        & \multicolumn{2}{c}{Block position}
        & \multicolumn{2}{c}{Block angle} \\
        \cmidrule(lr){2-3}\cmidrule(lr){4-5}\cmidrule(l){6-7}
        Representation & Linear & MLP & Linear & MLP & Linear & MLP \\
        \midrule
        \multicolumn{7}{@{}l}{\textit{Run 1}} \\
        Global embedding & 0.985 & 0.996 & 0.993 & 0.998 & 0.873 & 0.901 \\
        Concatenated slots & 0.999 & 0.999 & 0.999 & 0.999 & 0.941 & 0.952 \\
        \midrule
        Slot 0 & 0.732 & 0.811 & 0.950 & 0.975 & 0.680 & 0.831 \\
        Slot 1 & 0.522 & 0.968 & 0.763 & 0.991 & 0.424 & 0.923 \\
        Slot 2 & 0.682 & 0.694 & \textbf{0.999} & \textbf{0.999} & \textbf{0.938} & \textbf{0.978} \\
        Slot 3 & \textbf{0.999} & \textbf{1.000} & 0.794 & 0.846 & 0.474 & 0.578 \\
        \midrule
        \multicolumn{7}{@{}l}{\textit{Run 2}} \\
        Global embedding & 0.983 & 0.995 & 0.992 & 0.998 & 0.850 & 0.897 \\
        Concatenated slots & 0.999 & 0.999 & 0.999 & 1.000 & 0.939 & 0.963 \\
        \midrule
        Slot 0 & 0.676 & 0.966 & 0.884 & 0.986 & 0.397 & 0.897 \\
        Slot 1 & \textbf{0.999} & \textbf{0.999} & 0.873 & 0.917 & 0.526 & 0.669 \\
        Slot 2 & 0.915 & 0.967 & 0.963 & 0.984 & 0.677 & 0.839 \\
        Slot 3 & 0.688 & 0.708 & \textbf{0.999} & \textbf{0.999} & \textbf{0.934} & \textbf{0.976} \\
        \midrule
        \multicolumn{7}{@{}l}{\textit{Run 3}} \\
        Global embedding & 0.984 & 0.994 & 0.992 & 0.998 & 0.875 & 0.912 \\
        Concatenated slots & 0.997 & 0.999 & 0.999 & 1.000 & 0.927 & 0.960 \\
        \midrule
        Slot 0 & 0.880 & 0.946 & 0.959 & 0.984 & 0.702 & 0.860 \\
        Slot 1 & 0.908 & 0.985 & 0.930 & 0.980 & 0.491 & 0.819 \\
        Slot 2 & 0.803 & 0.876 & 0.994 & 0.997 & 0.860 & 0.925 \\
        Slot 3 & \textbf{0.996} & \textbf{0.999} & \textbf{0.998} & \textbf{1.000} & \textbf{0.908} & \textbf{0.964} \\
        \bottomrule
    \end{tabular}
\end{table}

\end{document}